\documentclass[letterpaper, 10 pt, conference]{IEEEtran}

\IEEEoverridecommandlockouts                              

\usepackage{cite}
\usepackage{amsmath,amssymb,amsfonts}
\usepackage{algorithmic}
\usepackage{graphicx}
\usepackage{textcomp}
\usepackage{hyperref}
\usepackage[table,xcdraw]{xcolor}
\usepackage{booktabs}
\usepackage{graphicx}
\usepackage[normalem]{ulem}
\useunder{\uline}{\ul}{}
\usepackage{float}
\usepackage{booktabs}
\usepackage[pscoord]{eso-pic}

\usepackage[numbers]{natbib}
\newcommand{\etal}{\textit{et~al.}}

\hypersetup{
    colorlinks=true,
    linkcolor=blue,
    filecolor=magenta,      
    urlcolor=blue,
    citecolor=blue,
    pdfpagemode=FullScreen,
    }

\def\BibTeX{{\rm B\kern-.05em{\sc i\kern-.025em b}\kern-.08em
    T\kern-.1667em\lower.7ex\hbox{E}\kern-.125emX}}
\begin{document}

\title{Contrastive Language, Action, and State Pre-training for Robot Learning}

\author{\IEEEauthorblockN{Krishan Rana*, Andrew Melnik and Niko S\"underhauf}
\thanks{Krishan Rana and Niko S\"underhauf are with the QUT Centre for Robotics, and Andrew Melnik is with the University of Bielefeld. We acknowledge continued support from the Queensland University of Technology (QUT) through the Centre for Robotics. This work was supported by an Australian Research Council Discovery Project (project number DP220102398). *Correspondence email: \texttt{ranak@qut.edu.au}}
}

\maketitle

\begin{abstract}
In this paper, we introduce a method for unifying language, action, and state information in a shared embedding space to facilitate a range of downstream tasks in robot learning.  Our method, Contrastive Language, Action, and State Pre-training (CLASP), extends the CLIP formulation by incorporating distributional learning, capturing the inherent complexities and one-to-many relationships in behaviour-text alignment. By employing distributional outputs for both text and behaviour encoders, our model effectively associates diverse textual commands with a single behaviour and vice-versa. We demonstrate the utility of our method for the following downstream tasks: zero-shot text-behaviour retrieval, captioning unseen robot behaviours, and learning a behaviour prior for language-conditioned reinforcement learning. Our distributional encoders exhibit superior retrieval and captioning performance on unseen datasets, and the ability to generate meaningful exploratory behaviours from textual commands, capturing the intricate relationships between language, action, and state. This work represents an initial step towards developing a unified pre-trained model for robotics, with the potential to generalise to a broad range of downstream tasks.
\end{abstract}


\section{Introduction}

Recent advancements in the fields of natural language processing and computer vision have demonstrated the potential of large-scale pre-trained models in learning general representations for a wide range of downstream tasks \cite{Radford2021LearningTV, OpenAI2023GPT4TR, Li2023BLIP2BL, Driess2023PaLMEAE}. However, the robotics community is yet to develop a unified representation that can encapsulate the rich and diverse information inherent in robotic systems, spanning language, action, and state. Such a unified representation could significantly improve the performance and generalization of robot learning algorithms, enabling seamless integration of language understanding and high-level task execution.

\begin{figure}[t]
  \centering
  \includegraphics[width=0.5\textwidth]{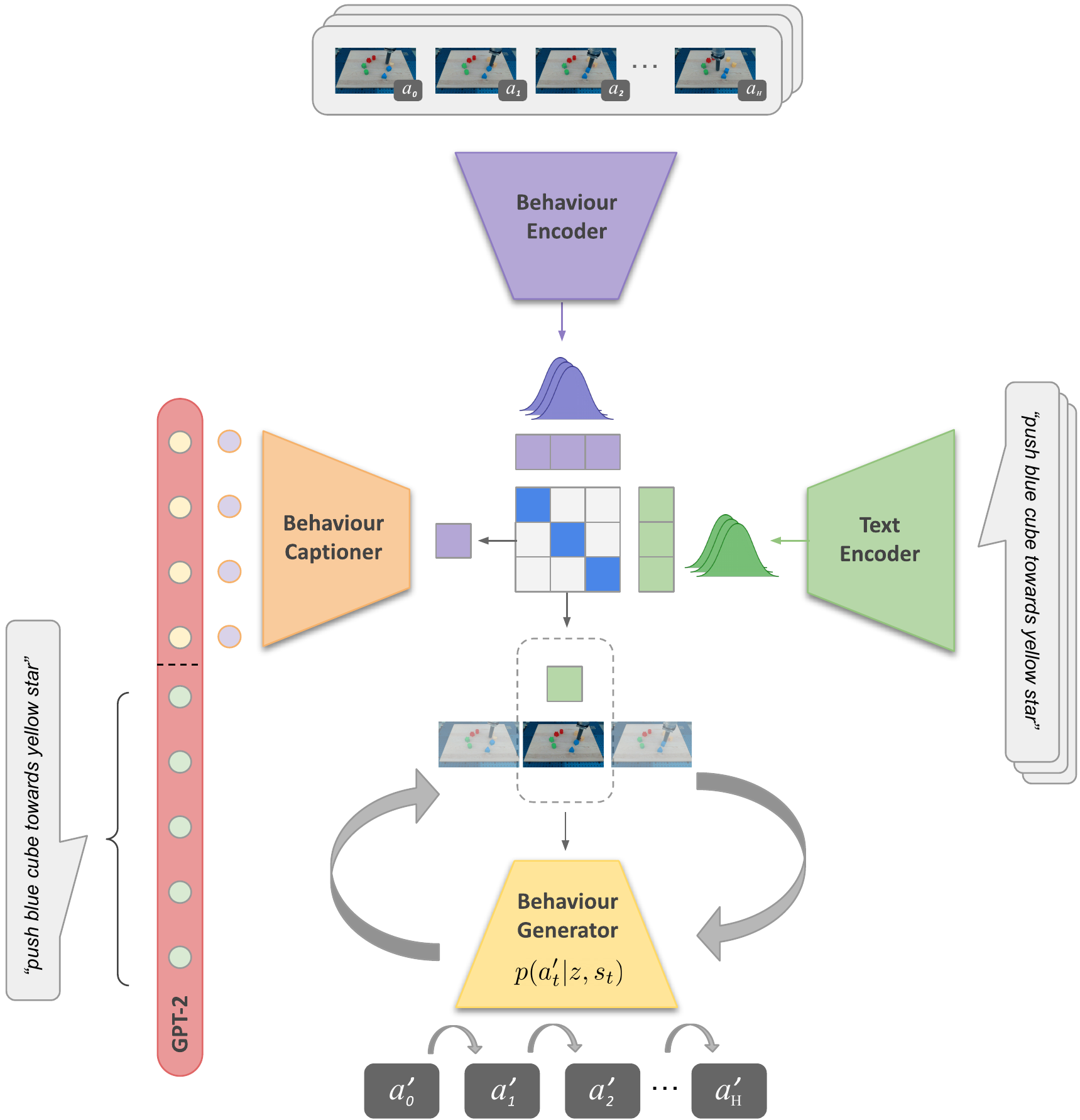}
  \caption{We propose distributional encoders for both robot behaviour trajectories and textual descriptions and align the embedding space using a contrastive loss via the reparameterisation trick. We additionally regularise this space using two additional loss terms which encourage the embeddings to better align with the desired downstream tasks for behaviour captioning and generation.}
  \label{clasp}
\end{figure}

Multi-modal contrastive models such as CLIP \cite{Radford2021LearningTV} have demonstrated the ability to develop these desired intricate relationships between text and images, by learning shared representations that facilitate a wide range of downstream tasks including text-to-image generation \cite{Ramesh2022HierarchicalTI}, image captioning \cite{Mokady2021ClipCapCP, Tang2021CLIP4CaptionCF}, image classification \cite{Radford2021LearningTV} and segmentation \cite{lueddecke22_cvpr, li2022languagedriven}. In this work, we explore what it takes to extend this idea to the robotics domain, where the alignment of language, states and actions can play a critical role in learning representations that can facilitate various downstream robot learning applications.

The direct application of the CLIP architecture to connecting language and behaviours in robotics raises several challenges. Unlike the image-text pairing in CLIP, where static images are matched with their corresponding descriptions, the robotics domain deals with continuous, dynamic sequences of state-action pairs and their varying textual representations, making the one-to-one mapping between modalities more complex. Inherently these modalities exhibit one-to-many relationships where a single textual command can correspond to multiple valid robot trajectories, and conversely, a single robot trajectory could be accurately described by multiple textual commands. This relationship demands an alternative learning approach that can capture the variability and nuances of such connections. 


In this paper, we adapt the standard CLIP architecture to the domain of robotics with the aim of establishing a unified embedding space for language, actions, and states through contrastive pre-training. To tackle the inherent bidirectional one-to-many mapping between robot behaviours and textual descriptions, we propose to model the encoders as distributions from which we sample via reparameterisation \cite{Kingma2013AutoEncodingVB}. The sampled embeddings are then aligned using the symmetric cross-entropy contrastive loss employed by CLIP. Furthermore, we encourage the model to learn generalizable and useful representations by regularizing the embedding space with two interconnected auxiliary downstream tasks: behaviour captioning and behaviour reconstruction.

Our proposed approach, known as CLASP (Contrastive Language, Actions, and State Pretraining), demonstrates superior retrieval performance compared to the traditional CLIP formulation when extended to the behaviour-text setting. Preliminary results additionally indicate the potential of the shared representation to facilitate downstream applications including robot behaviour captioning, text-conditioned behaviour generation and learning behaviour priors for reinforcement learning.



\section{Related Work}


In recent years, the pursuit of a shared representation for language, states, and actions has garnered significant interest in the robotics research community, aiming to develop more intuitive and versatile robotic systems. This endeavor involves the grounding of language in robotic actions and states, with various related works exploring different methodologies to achieve this goal.

\subsubsection{Language-Conditioned Robot Learning}
One prominent approach to grounding natural language in robot states and actions involves conditioning robot learning policies on language instructions \cite{Shridhar2022PerceiverActorAM, lynch2022interactive, Mees2022WhatMI, lynch2021language, carta2023grounding}. This approach is based on the premise that an effective connection between language, actions, and states can be established via the task-centric loss function. However, applying these techniques in isolation frequently results in overfitting to initial object scenes, yielding suboptimal embeddings and limited task generalisation \cite{Jang2022BCZZT}. To mitigate these issues, prior research has explored the use of auxiliary multi-modal alignment losses \cite{Jang2022BCZZT, Abramson2021CreatingMI, Mees2022WhatMI, Hermann2017GroundedLL} in conjunction with standard imitation or reinforcement learning objectives. These studies have demonstrated that integrating appropriate regularisation with these objectives fosters the development of a more structured and coherent embedding space \cite{Jang2022BCZZT}, substantially accelerating learning and improving the generalisation capabilities of the learned policy. In this work, we examine the multi-modal model alignment module in greater detail and investigate the requirements for learning an effective representation that can facilitate a variety of robot learning applications.

\subsubsection{Pretraining for Robot Learning} 
In the domain of pretraining for robot learning, several studies have focused on the development of versatile shared embedding spaces using large-scale pre-training on video and language data. Nair \etal \cite{Nair2021LearningLR} use this to learn a general visual representation that could be used across a wide range of robot learning tasks. Fan \etal \cite{fan2022minedojo} build on this idea to train a similar shared embedding space and utilise the cosine similarity between text and video embeddings as a reward for downstream RL. Xiao \etal \cite{Xiao2022RoboticSA} finetune the CLIP model to align start-end images of a robot trajectory with textual descriptions in order to relabel new datasets. All these ideas demonstrate the versatility of a shared embedding space for different components of the robot learning pipeline. In our work, we extend these ideas to include robot actions within the multi-modal embedding space and explore the applicability of the shared representation to facilitate other downstream tasks including behaviour captioning as well as behaviour generation.

\section{Motivation}


The growing interest in developing shared representations across various modalities, such as text, images, and audio, has led to significant advancements in natural language processing and computer vision \cite{Radford2021LearningTV, guzhov2021audioclip, Li2023BLIP2BL}. However, the robotics domain, which encompasses language, states, and actions, has yet to witness a dedicated effort to create a unified representation that can facilitate more natural and versatile robotic systems. The potential benefits of such a representation in robotics are immense, including improved performance and generalisation of robotic learning algorithms and seamless integration of language understanding with high-level task execution. Although some works have indirectly addressed this challenge, a focused approach to establishing a shared representation for the complex robotics domain is still lacking. Unique challenges arise from the continuous and dynamic nature of state-action pairs, the diverse textual representations, and the inherent one-to-many relationships between language and robot behaviours. To tackle these complexities, innovative learning approaches capable of effectively capturing the intricacies of the relationships between language, actions, and states are needed, thus motivating the pursuit of novel methodologies in this research area.

\section{Problem Formulation}

In this section, we formally define the problem of learning a shared embedding space for language, actions and states in the context of robotics. 

Let $\mathcal{L}$ denote the language modality, where each element $l \in \mathcal{L}$ represents a natural language description or command. Similarly, let $\mathcal{B}$ denote the behavior modality, where each element $b \in \mathcal{B}$ represents a robot behavior, consisting of a sequence of state-action pairs $(s_1, a_1, s_2, \dots, s_T, a_T)$, with $T$ denoting the length of the sequence. The objective is to learn a shared embedding space $\mathcal{Z}$, where the elements $z_l \in \mathcal{Z}$ and $z_b \in \mathcal{Z}$ correspond to the embeddings of language and behaviours, respectively. This space should capture the complex relationships between these modalities and enable efficient transfer learning across various robot learning tasks. To achieve this, we propose a multi-modal contrastive learning framework, which consists of two encoder networks, $\phi_l: \mathcal{L} \rightarrow \mathcal{Z}$ and $\phi_b: \mathcal{B} \rightarrow \mathcal{Z}$. These encoders are designed to project the elements of each modality into the shared embedding space $\mathcal{Z}$, preserving the rich relationships between language and behaviours. The learning objective is to minimize a contrastive loss function that encourages the alignment of corresponding language and behaviour embeddings while pushing apart non-matching pairs. The proposed framework should also take into account the inherent bidirectional one-to-many mappings between robot behaviours and textual descriptions, as well as the temporal dependencies between state-action sequences.

\section{Methodology}

In this section, we present our methodology for learning a shared embedding space for language and behaviours in the context of robotics. We leverage contrastive learning, with distributional encoders to facilitate the learning of one-to-many mappings between text and behaviours and vice versa.  Additionally, we incorporate two auxiliary tasks for regularising the model and improving generalisation. Full implementation details are provided in the Appendix section.

\subsection{Distributional Encoders and Sampling}

We utilise two distributional encoders, one for language ($\phi_l$) and one for behaviours ($\phi_b$), that output the parameters of a Gaussian distribution in the shared embedding space. For each language description $l$ and behaviour $b$, we compute the mean ($\mu$) and variance ($\sigma^2$) of the corresponding embeddings using the respective encoders:

\begin{equation}
\begin{aligned}
\mu_l, \sigma^2_l &= \phi_{l}(l), \hspace{1.5cm}
\mu_b, \sigma^2_b &= \phi_{b}(b)
\end{aligned}
\end{equation}

To obtain the required embeddings for alignment, we sample from these distributions using the reparameterisation trick \cite{Kingma2013AutoEncodingVB}, commonly used when training variational auto-encoder networks: 

\begin{equation}
\begin{aligned}
z_b &= \mu_b + \epsilon_b \odot \sqrt{\sigma^2_b}, \hspace{1cm} z_l &= \mu_l + \epsilon_l \odot \sqrt{\sigma^2_l}
\end{aligned}
\end{equation}

\noindent where $\epsilon_b$ and $\epsilon_l$ are random noise vectors drawn from a standard normal distribution ($\epsilon \sim \mathcal{N}(0, 1)$) and $\odot$ denotes element-wise multiplication.

\subsection{Behaviour-Language Alignment Loss}
To align the embeddings of behaviours and their corresponding textual descriptions, we utilise the same contrastive objective used for pairing images and captions in CLIP \cite{Radford2021LearningTV} which is based on the symmetric cross-entropy loss. Given a mini-batch of $N$ samples, the loss function for our behaviour-text alignment is given by:

\begin{multline}
\mathcal{L}_{\text{align}} = -\frac{1}{2N}\sum_{i=1}^{N}\Bigg[\log \frac{\exp{\left\langle z_{b_i}, z_{l_i} \right\rangle / \tau}}{\sum_{j=1}^{N} \exp{\left\langle z_{b_i}, z_{l_j} \right\rangle / \tau}} + \\
\log \frac{\exp{\left\langle z_{l_i}, z_{b_i} \right\rangle / \tau}}{\sum_{j=1}^{N} \exp{\left\langle z_{l_i}, z_{b_j} \right\rangle / \tau}}\Bigg]
\end{multline}

Here, $\left\langle \cdot, \cdot \right\rangle$ denotes the inner product between two vectors, $z_{b_i}$ is the behaviour embedding for the $i$-th sample, $z_{t_i}$ is the corresponding text embedding, and $\tau$ is a temperature hyperparameter. This loss encourages the model to align behaviour and text embeddings for each sample while pushing them away from the embeddings of other samples in the batch. Similar to the original CLIP model, the loss is computed for both behaviour-to-text and text-to-behaviour directions, and their average is used for optimization.

\subsection{Auxiliary Tasks}

To regularise the model and improve generalisation, we introduce two interconnected auxiliary tasks: behaviour captioning and behaviour generation.

\subsubsection{Behaviour Captioning}

The goal of this task is to predict the natural language description $l$ of a given behaviour sequence $b$. Following from the CLIPCap model presented by Mokady \etal \cite{Mokady2021ClipCapCP}, we assume that all the necessary information for captioning a behaviour $b$ is present in the sampled embedding $z_b$, given its alignment with text in the shared embedding space.  This in essence should allow us to predict the corresponding caption directly from the behaviour embedding:

\begin{equation}
\max _\theta \sum_{i=1}^N \log p_\theta\left(l_1^i, \ldots, l_{n}^i \mid z_b^i\right),
\end{equation}

\noindent where we refer to the captions as a sequence of tokens $l=l_1, \ldots, l_{n}$ padded to a maximum length $n$. Similar to \cite{Mokady2021ClipCapCP}, we focus on prefix fine-tuning \cite{Li2021PrefixTuningOC} as a sample efficient strategy for training our captioning network. We utilise a pre-trained GPT-2 \cite{Radford2019LanguageMA} model as the backbone of the captioning network and solely train a mapping network $\psi$ which projects our behaviour embedding $z_b$ to $k$ embedding vectors suitable as input to the large language model:

\begin{equation}
p_1^i, \ldots, p_k^i=\psi\left(z_b\right)
\end{equation}

The final objective for training
the mapping component $\psi$ is to predict the caption tokens conditioned on the prefix in an auto-regressive fashion using the cross-entropy loss:

\begin{equation}
\mathcal{L}_{\text{caption}}=-\sum_{i=1}^N \sum_{j=1}^{n} \log p_\theta\left(l_j^i \mid p_1^i, \ldots, p_k^i, l_1^i, \ldots, l_{j-1}^i\right)
\end{equation}

\subsubsection{Behavior Generation}

In the behaviour generation task, we aim to reconstruct the encoded action sequence $\textbf{\textit{a}}$ that is processed by the behaviour encoder from the sampled language embedding $z_l$. Given that robot environments can be dynamic, we model the behaviour generator $\pi$ as a closed-loop policy that conditions on both the text embedding and the current state $s_t$ in order to generate the corresponding action $a'_t$. This network is trained using the mean squared error (MSE) loss, which measures the difference between the predicted action sequence and the ground truth action sequence:

\begin{equation}
L_{\pi} = \frac{1}{N} \sum_{n=1}^{N} \frac{1}{T} \sum_{t=1}^{T} \left||a_t - \pi(a'_{t} | z_l, s_{t})|\right|_2^2,
\end{equation}

\noindent where $a_t$ denotes the $t$-th ground truth action in the sequence, $\pi(a'_{t} | z_l, s{t})$ denotes the predicted action given the text embedding $z_l$ and state $s_t$, and $T$ is the total number of actions in the sequence.

\subsection{Total Loss}

The total loss for our model is a combination of the three loss terms, weighted by their respective hyperparameters $\beta$:

\begin{equation}
    \mathcal{L}_{\text{CLASP}} = \beta_1\mathcal{L}_{\text{align}} + \beta_2\mathcal{L}_{\text{caption}} + \beta_3\mathcal{L}_{\pi}
\end{equation}

Our model is trained by minimizing this loss, which encourages the learned shared embedding space to structure itself such that it can effectively align language, actions, and states. By striking a balance between these objectives, the model learns to capture the intricacies and nuances of the relationships between these modalities, leading to better generalisation and performance on downstream applications.

\section{Evaluation}

\subsection{Training Dataset}

We train and evaluate our method on the \texttt{Language-Table} demonstration dataset and environment \cite{lynch2022interactive} which consists of a diverse range of robot state-action trajectories coupled with natural language descriptions of the corresponding behaviour. The dataset consists of both real-world and simulation demonstration data. The environment consists of a xArm5 robot, constrained to move in a 2D plane, with a cylindrical end-effector, in front of a smooth wooden board with a fixed set of 8 plastic blocks, comprising 4 colours and 6 shapes. Actions are 2D delta Cartesian setpoints, from the previous setpoint to the new one. State information consists of RGB third-person images of the robot and board as shown in Figure \ref{res:caption}.

\subsection{Alignment Evaluation}

\begin{table}[!t]
\caption{Zero-shot retrieval on an unseen dataset}
\label{tab:retrieval}
\resizebox{\columnwidth}{!}{%
\begin{tabular}{@{}ccccccccl@{}}
\cmidrule(r){1-8}
\multicolumn{4}{c}{\textbf{Text Retrieval}}                                                                                                                              & \multicolumn{4}{c}{\textbf{Behaviour Retrieval}}                                                                                                                        &  \\
\multicolumn{2}{c}{\cellcolor[HTML]{FFFFFF}CLASP}         & \multicolumn{2}{c}{\cellcolor[HTML]{EFEFEF}\begin{tabular}[c]{@{}c@{}}CLASP\\ (Distributional)\end{tabular}} & \multicolumn{2}{c}{\cellcolor[HTML]{FFFFFF}CLASP}         & \multicolumn{2}{c}{\cellcolor[HTML]{EFEFEF}\begin{tabular}[c]{@{}c@{}}CLASP\\ (Distributional)\end{tabular}} &  \\
\cellcolor[HTML]{FFFFFF}R@1 & \cellcolor[HTML]{FFFFFF}R@5 & \cellcolor[HTML]{EFEFEF}R@1                           & \cellcolor[HTML]{EFEFEF}R@5                          & \cellcolor[HTML]{FFFFFF}R@1 & \cellcolor[HTML]{FFFFFF}R@5 & \cellcolor[HTML]{EFEFEF}R@1                           & \cellcolor[HTML]{EFEFEF}R@5                          &  \\ \cmidrule(r){1-8}
\cellcolor[HTML]{FFFFFF} 40.0   & \cellcolor[HTML]{FFFFFF}  73.3  & \cellcolor[HTML]{EFEFEF}             \textbf{73.3}                 & \cellcolor[HTML]{EFEFEF}                     \textbf{ 93.3}       & \cellcolor[HTML]{FFFFFF} \textbf{86.7}   & \cellcolor[HTML]{FFFFFF} 100   & \cellcolor[HTML]{EFEFEF}             66.7               & \cellcolor[HTML]{EFEFEF}                    \textbf{100 }        &  \\ \cmidrule(r){1-8}
\end{tabular}%
}
\end{table}

We evaluate the zero-shot retrieval accuracy of our model against a non-distributional variant on a held-out dataset from the \texttt{Language-Table} suite, comparing their top-1 and top-5 retrieval accuracies for both text and behaviour. We summarise the results in Table \ref{tab:retrieval}.

Our distributional encoders demonstrate improved generalization, with a 33.3\% increase in top-1 text retrieval accuracy and a 20.0\% increase in top-5 accuracy. Although top-5 behaviour retrieval accuracy remains at 100\% for both variants, the distributional approach shows a drop in top-1 accuracy.

This discrepancy could be due to the distributional encoders capturing a broader range of behaviour representations, which enhances text retrieval performance but makes it harder to pinpoint the exact behaviour in top-1 results. Nevertheless, the distributional variant effectively identifies relevant behaviours within the top-5 results, suggesting that it is sensitive to subtle behaviour differences and could excel in fine-grained retrieval tasks. Further investigation is needed to confirm this hypothesis and understand the trade-offs between the distributional and non-distributional CLASP variants.





\subsection{Behaviour Captioning}

\begin{table}[!t]
\centering
\caption{Behaviour-To-Description Translation Accuracy}
\label{tab:caption}
\resizebox{0.8\columnwidth}{!}{%
\begin{tabular}{@{}cccc@{}}
\toprule
\textbf{CLASP (Distributional)} & \textbf{CLASP} & \textbf{Seq2Seq}\\ \midrule
           \textbf{ 46.6\%  }             &       40.0\%                         &      26.6\%             \\ \bottomrule
\end{tabular}%
}
\end{table}

We further evaluate our model's ability to caption unseen behaviour trajectories by comparing its performance with a non-distributional variant (CLASP) and a Seq2Seq baseline that does not utilise the aligned representation space for captioning. The results are summarised in Table \ref{tab:caption}.

Our model achieves a 46.6\% translation accuracy, surpassing both alternatives. This improved performance can be attributed to the aligned embedding space, which enables effective information transfer between behaviour sequences and text captions, leading to increased captioning accuracy. The Seq2Seq baseline, without the shared space, only attains 26.6\% accuracy, while the non-distributional CLASP variant reaches 40.0\% accuracy. These findings suggest that incorporating distributional encoders allows our model to capture the nuanced behaviour-text relationships, resulting in better captioning performance.

We present qualitative captioning examples of our approach in Figure \ref{res:caption}. Additionally, we provide the ground truth language descriptions from the dataset to emphasise the one-to-many mapping nature of behaviours-to-textual descriptions.

It is worth noting that a significant number of captioning failures stem from the visual model's inability to distinguish subtle object differences in our dataset, such as "yellow pentagon" from "yellow star" or "green cube" from "green star." In our experiments, we used a frozen CLIP visual encoder to process these images, which is not specifically designed for fine-grained object-level feature extraction. We believe that using a fine-tuned or alternative model could yield better results.

\subsection{Behaviour Generation}

In our final evaluation, we investigate the effectiveness of the shared embedding space for meaningful behaviour generation. The ability to generate useful behaviours from either a sampling space or textual descriptions is crucial for various robotic learning applications, such as facilitating exploration in reinforcement learning \cite{Rana2022ResidualSP, singh2020parrot, pertsch2020spirl}, model predictive control \cite{shi2022skimo}, or dataset generation for imitation learning. We assess the behaviour generator's capacity to produce meaningful behaviours in the context of the \texttt{Language-Table} environment. Here, useful behaviours are defined as trajectory sequences that result in block rearrangements constrained to the board region. We evaluate our skill generator based on this criterion and compare its performance with the standard random exploration approach. We additionally learn a state-conditioned behaviour prior over this embedding space using the approach proposed in \cite{Rana2022ResidualSP}. The results, presented in Table \ref{tab:behaviour_gen}, demonstrate that both methods leveraging the distributional embedding space induced by CLASP can produce a high proportion of useful behaviours in the environment, outperforming random exploration alone.


\begin{table}[!t]
\caption{Percentage of Useful Trajectories Generated During Exploration}
\label{tab:behaviour_gen}
\resizebox{\columnwidth}{!}{%
\begin{tabular}{@{}l
>{\columncolor[HTML]{EFEFEF}}c 
>{\columncolor[HTML]{EFEFEF}}l l@{}}
\toprule
                             & \multicolumn{2}{c}{\cellcolor[HTML]{EFEFEF}\textbf{CLASP}}                                                &                             \\ \midrule
\textbf{Method}              & \multicolumn{1}{l}{\cellcolor[HTML]{EFEFEF}\textbf{Behaviour Prior}} & \textbf{Text Encoding}                           & \textbf{Random Exploration} \\ \midrule
\textbf{Useful Trajectories} & 87.7\%                                                              & \multicolumn{1}{c}{\cellcolor[HTML]{EFEFEF}60.0\%} & \multicolumn{1}{c}{27.7\%}     \\ \bottomrule
\end{tabular}%
}
\end{table}

\section{Conclusions}

This body of work represents an initial step towards developing a shared embedding space for language, states, and actions in robotics. Preliminary results indicate that accounting for the bidirectional one-to-many nature of the text-behaviour relationship is essential when constructing this shared representation, especially for downstream tasks involving generative modeling. Our approach demonstrates improved retrieval performance compared to non-distributional methods and showcases the applicability of the shared embedding space across two distinct downstream tasks. It is important to note that the evaluation conducted in this study focused on a single robot domain, and further assessments across larger and more diverse datasets are necessary to establish the viability of the approach. We encourage continued research in this area as more extensive and varied datasets become available to the robot learning community.


\section*{Acknowledgments}

 The authors would like to thank Jad Abou-Chakra, Sourav Garg, Mingda Xu, Jesse Haviland and Robert Lee for their valuable and insightful discussions towards this contribution.

\bibliography{manual2.bib}

\newpage

\appendix

\begin{figure*}[htbp]
  \includegraphics[width=\textwidth]{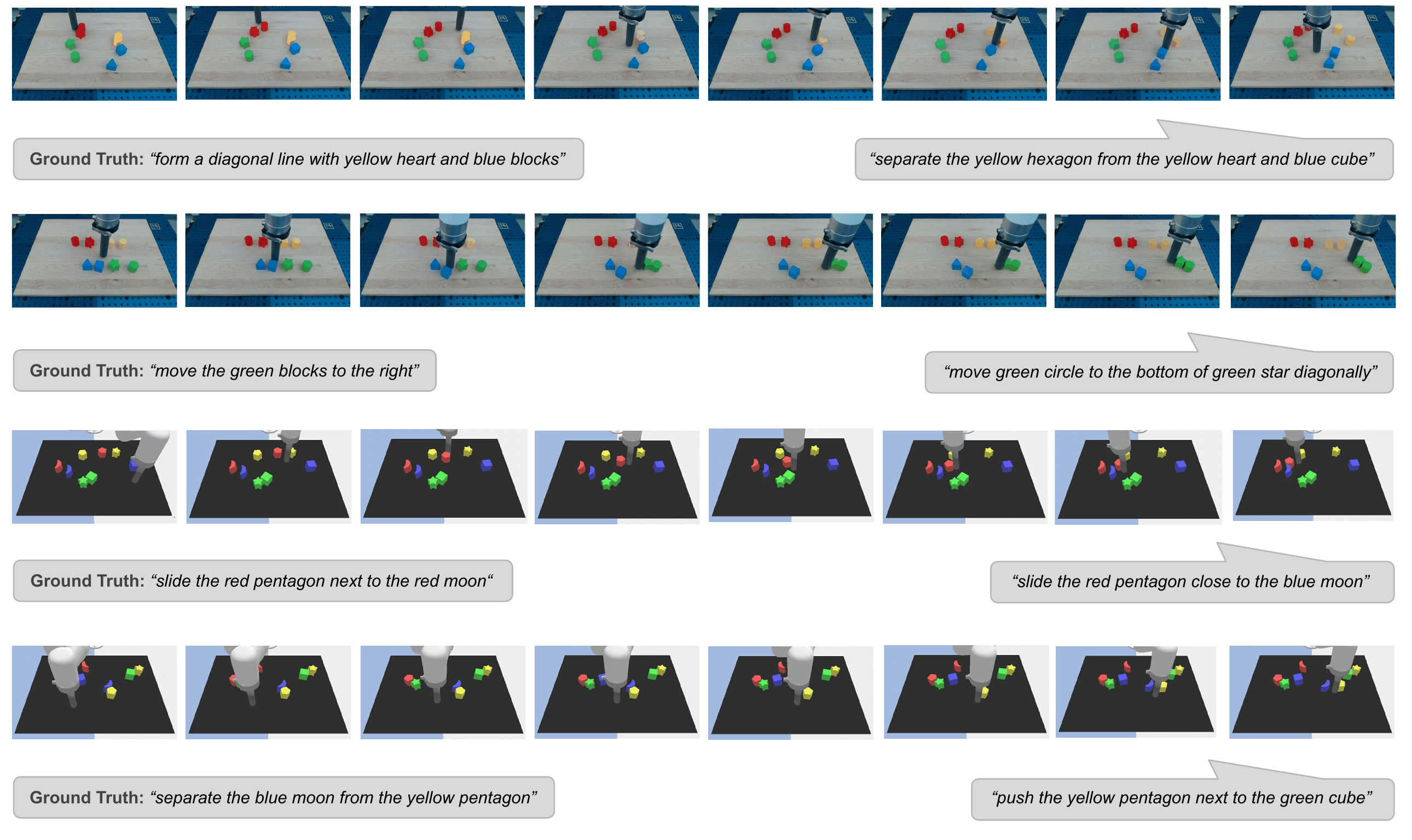}
  \caption{Captioning unseen robot trajectories using our trained model on both real and simulation environments. Note the significant discrepancy between the labelled ground truth description and the generated description. This highlights the one-to-many relationship between the behaviours and textual descriptions.}
  \label{res:caption}
\end{figure*}

\section{Implementation Details}

\subsection{Behaviour Encoder}

Each training example consists of $(s, a, l) \sim D$, where $s \in \mathbb{R}^{T \times 320 \times 180 \times 3}$ is the RGB observation history with varying sequence length $T$ for the behaviour trajectory, $a \in \mathbb{R}^2$ is the delta Cartesian action of the robot arm, and $l$ is the natural language instruction carried out by the robot. Each image frame is passed through a pre-trained CLIP image encoder to obtain a visual feature representation $f_s \in \mathbb{R}^{512}$ before concatenating it with the normalized robot action. We prepend a \texttt{[CLS]} token to this sequence, which will later serve as the final representation for the behaviour sequence. The sequence of state-action pairs, including the \texttt{[CLS]} token, is processed by an MLP to obtain the shape $[T+1, \texttt{dmodel}]$, and 2D position-encoded before being passed through a standard transformer encoder model \cite{vaswani_transformer}. Our behaviour encoder transformer has 2 layers, with $\texttt{dmodel} = 512$, 2 heads, a feed-forward width of 128, and a dropout rate of 0.1. The \texttt{[CLS]} token is then processed by a 3-layer MLP, which outputs the desired mean and sigma for the encoder distribution, both of dimension 512.

\subsection{Text Encoder}

We employ a pre-trained CLIP text encoder \cite{Radford2021LearningTV} for processing the input instruction text. The text is preprocessed by removing punctuation and extra spaces and is fed into a pre-trained CLIP text encoder, which produces a textual feature representation $f_l \in \mathbb{R}^{512}$. Subsequently, the textual representation is passed through a 3-layer MLP projector head, which generates the 512-dimensional mean and sigma for the desired distributional output of the textual encoder.

\subsection{Behaviour Generator}
The role of the behaviour generator is to take an embedding $z$ from the shared latent space and map it to a sequence of meaningful actions captured by the behaviour dataset. Due to the dynamic nature of robot environments, we make this generation process closed-loop by conditioning it on the current state. As a result, the behaviour generator can be viewed as a closed-loop policy:

\begin{equation}
a_t = \pi(z, s_t)
\end{equation}

We model the policy $\pi$ as a 6-layer MLP, which takes the sampled embedding $z$ and the current state $s_t$ as inputs and outputs the corresponding 2-dimensional action $a_t$. The embedding space encapsulates a diverse range of behaviours, and selecting an appropriate $z$ for sampling can be challenging. In this work, we describe two strategies used for evaluation:

\subsubsection{Text-Conditioned Generation}

In this approach, we leverage the shared embedding space and distributional nature of our encoders to sample from the embedding space during inference. Given the alignment between text and behaviours, we can map textual commands to a distribution over $z$, from which we can sample an appropriate $z$ for decoding into a behaviour sequence. By doing so, we effectively utilize the learned connections between language and behaviour to generate meaningful action sequences based on the input textual commands.

\subsubsection{State-Conditioned Behaviour Prior}

We additionally explore the ability to utilise this shared embedding space to learn a state-conditioned behaviour prior. In this case the prior is task agnostic and captures the entire range of state-relevant behaviours in the embedding space. Such a prior has been shown to be useful for accelerating RL exploration \cite{pertsch2020spirl, Rana2022ResidualSP}. We follow the same strategy used in \cite{Rana2022ResidualSP} to learn a state conditioned prior over an existing embedding space using normalising flows. The network parameterising the behaviour prior $f:\mathcal{Z}\times \mathcal{S} \rightarrow G$ is a conditional real NVP \cite{dinh2016density} which consists of four affine coupling layers, where each coupling layer takes as input the output of the previous coupling layer, and the robot state vector $s_0$ from the start of the behaviour sequence. We use a standard Gaussian $p_{\mathcal{G}} (g)\sim\mathcal{N}(0,I)$ as our base distribution for our generative model. We refer the reader to \cite{Rana2022ResidualSP} for a more detailed treatment of this model. The loss function for a single example is given by:
\begin{equation}
    \mathcal{L}_{prior} = \log p_{\mathcal{G}}(f(z, s_0)) + \log \left| \det \frac{\partial f}{\partial z^\top} \right|.
\end{equation}
Once trained, the flow model allows us to sample an appropriate $z$ from the embedding space via the bijective mapping function $f^{-1}(g, s)\sim p(z|s)$, where $g$ is sampled from a simple Gaussian distrbution ${\mathcal{N}}(0,I)$.

\subsection{Behaviour Captioner}

As previously mentioned, the captioning network comprises a trainable mapping network and a frozen GPT-2 decoder network. The mapping network's purpose is to project a sampled behaviour embedding $z_b$ into $K$ token embeddings, which can then be passed as input to the GPT model. Our mapping network consists of 8 multi-head self-attention layers, each with 8 heads. We set the prefix length $K$ to 10. Once trained, the decoding process is performed through beam search to obtain the natural language description of a behaviour. Considering the one-to-many mapping from behaviours to text, we evaluate captioning performance manually via visual inspection, assessing the quality and relevance of the generated captions to the behaviour video stream.

\end{document}